\documentclass[letterpaper,10pt,conference]{ieeeconf}
\IEEEoverridecommandlockouts
\usepackage{amsmath}
\usepackage{amssymb}
\usepackage{booktabs}
\usepackage{multirow}
\usepackage{graphicx}
\usepackage{cite}
\usepackage{url}
\usepackage{float}

\title{\LARGE\bf EgoPriMo: Egocentric Motion Generation for Interactive Humanoid Control}
\author{Haoyang Ge$^{1,2,\ast}$, Peng Ren$^{2,3}$, Yukun Shi$^{2,4}$,
Cong Huang$^{2,4}$, Kun Li$^{1,\dagger}$, and Kai Chen$^{2,4,5,\dagger}$%
\thanks{$^{1}$Tianjin University, Tianjin, China.}%
\thanks{$^{2}$Zhongguancun Academy, Beijing, China.}%
\thanks{$^{3}$Beihang University, Beijing, China.}%
\thanks{$^{4}$Zhongguancun Institute of Artificial Intelligence, Beijing, China.}%
\thanks{$^{5}$DeepCybo, Beijing, China.}%
\thanks{$^{\ast}$\texttt{ghy0623@tju.edu.cn}.}%
\thanks{$^{\dagger}$Corresponding authors: Kun Li and Kai Chen.}%
}

\newcommand{\method}{EgoPriMo}
\newcommand{\methodfull}{Egocentric Motion Prior for Humanoid Robots}
\newcommand{\vx}{\mathbf{x}}
\newcommand{\vy}{\mathbf{y}}

\IEEEaftertitletext{
\begin{minipage}{\textwidth}
  \centering
  \includegraphics[width=\linewidth]{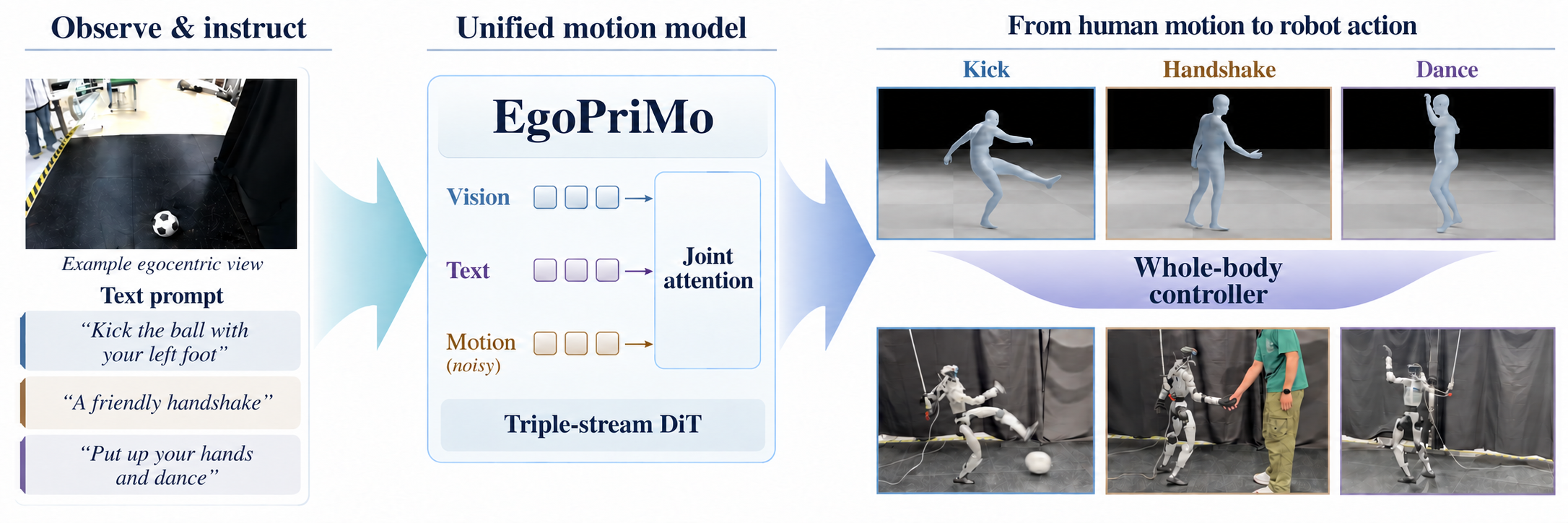}
  \makeatletter
  \def\@captype{figure}
  \makeatother
  \caption{\textbf{\method{}.}
Egocentric observations provide scene context, while text prompts specify
the intended action. Conditioned on both, \method{} generates SMPL
whole-body motion references for an existing controller to track,
enabling humanoid behaviors such as kicking, handshaking, and dancing.}
  \label{fig:head}
\end{minipage}%
\vspace{1em}}

\begin{document}
\maketitle
\thispagestyle{empty}
\pagestyle{empty}

\begin{abstract}
Humanoid robots require whole-body motions that adapt to scene context, task
requirements, and user intent. Despite advances in motion tracking and
humanoid vision-language-action models, integrating scene context and user
intent into a reusable motion prior remains challenging.
We introduce \method{}, short for \methodfull{}, a unified framework
that learns such priors from egocentric human demonstrations. Given
egocentric observations and a text prompt, \method{} reconstructs, generates,
and forecasts SMPL-based full-body motion. Language is used as a high-level
control signal rather than a complete motion specification. At the core of
\method{} is a Triple-stream DiT that jointly models body dynamics, egocentric
visual context, and text; task-conditioning masks route different tasks and
missing-modality data through the same checkpoint. Experiments on Nymeria and
EgoExo4D demonstrate improved egocentric motion generation over UniEgoMotion,
with gains in motion accuracy, semantic similarity, and foot-contact quality.
Text prompts steer motion generation under fixed visual conditions, and the
generated SMPL motions can be executed through an existing whole-body
controller.
\end{abstract}


\section{Introduction}

Humanoid robots need natural, diverse, and controllable whole-body motions in
open environments. 
A practical humanoid should coordinate whole-body movements according to scene context and user intent. 
Examples include kicking a ball with the foot, reaching out for a handshake, and raising both hands to dance.
We study egocentric-observation conditioned full-body motion
generation as a scalable way to learn reusable motion priors from human
demonstrations.
As shown in Figure~\ref{fig:head}, \method{} translates egocentric observations
and text prompts into full-body motion references for humanoid execution.

Motion tracking and imitation have made strong progress in humanoid whole-body
control by reproducing trajectories from motion capture, teleoperation, or
curated libraries~\cite{humanplus2024, omnih2o2024, exbody2024, hover2024,
asap2025, visualmimic2025, sonic2025}. Yet the desired motion is usually already
specified, making these systems closer to motion replay than open-ended
synthesis. Humanoid vision-language-action models provide semantic instruction
interfaces~\cite{openvla2024, egovla2025, egomi2025}, but are still centered
on task-specific navigation, manipulation, or short-horizon operational skills.
Broad, long-horizon whole-body motion generation remains underexplored.
Egocentric human demonstrations bridge this gap. Egocentric videos are
abundant, low-cost, and capture visual context, camera-motion cues,
task-relevant objects, intent, and interaction cues. Recent egocentric datasets
and models demonstrate their value for activity understanding and human motion
modeling~\cite{grauman2024egoexo4d, nymeria2024, li2023egoego,
akada2024uniegomotion}. For humanoids, such data provide environment-grounded
supervision for learning both full-body motion and its coupling to the observed
scene.
A useful egocentric motion prior should reconstruct demonstrations, generate
full-body motion from partial egocentric evidence, and forecast future motion.
We therefore treat language as a high-level control signal rather than a
complete motion specification: users steer egocentrically grounded motion with
targeted prompts about intent, timing, or interaction style.

Egocentric observations and language constrain motion in complementary ways.
An image provides evidence about the surroundings but often leaves the
wearer's body outside the field of view, while a text prompt specifies the
intended action without prescribing a complete pose sequence. The model must
therefore combine scene evidence and action intent to generate temporally
coherent motion. This involves coordinating modalities with distinct roles:
visual features describe the observed context, text features encode semantic
intent, and motion features represent the evolving body state. These
considerations motivate a generative architecture that preserves
modality-specific processing while enabling joint cross-modal interaction.

As an egocentric motion prior for humanoid robots, \method{} unifies
reconstruction, generation, and forecasting. Given egocentric observations and a
text prompt, \method{} generates an SMPL-based full-body
motion sequence. At the core of \method{} is a Triple-stream DiT that jointly
models body motion, egocentric visual context, and language. Task-conditioning
masks decide which motion frames, image tokens, auxiliary pose cues, and text
tokens are visible, allowing the same checkpoint to support multiple tasks
without task-specific heads. The generated SMPL motion can be consumed by an
SMPL-conditioned humanoid controller.

Experiments on Nymeria show that one \method{} checkpoint supports
reconstruction, generation, and forecasting. On an external egocentric
generation benchmark, \method{} improves the strongest matched comparison in
joint accuracy and contact plausibility. These results suggest that egocentric
human motion generation is a practical route toward generalizable and
interactively controllable humanoid motion priors.
We further evaluate prompt control with 100 paired Nymeria validation clips,
holding egocentric conditions and initial noise fixed while changing the text.
The resulting motions shift toward the reference associated with the supplied
prompt. On a Unitree humanoid, the complete pipeline succeeds in 7 of 10
ball-kicking trials and 8 of 10 table-avoidance trials without robot-view
fine-tuning. Here, interaction means selecting a high-level prompt before
generation; online language correction and closed-loop replanning are outside
the evaluated setting.

Our main contributions are:
\begin{itemize}
    \item We formulate egocentric human motion generation as a scalable paradigm
    for humanoid full-body behavior learning.

    \item We introduce a Triple-stream DiT as the core generative backbone for
    jointly modeling body motion, egocentric visual context, and language.

    \item We propose a unified task-conditioned framework that performs
    full-body motion reconstruction, generation, and forecasting with a single
    checkpoint.
\end{itemize}

\section{Related Work}
\label{sec:related}

\textbf{Humanoid motion tracking and imitation.}
Motion imitation transfers human motion references to humanoid robots while
addressing embodiment and dynamics differences, as demonstrated by
ExBody~\cite{exbody2024} and ASAP~\cite{asap2025}. HumanPlus~\cite{humanplus2024}
and OmniH2O~\cite{omnih2o2024} extend motion tracking to teleoperation and
autonomous skill learning. HOVER~\cite{hover2024} unifies multiple control modes,
while SONIC~\cite{sonic2025} scales motion tracking and connects it to kinematic
planning and visuomotor policies. VisualMimic~\cite{visualmimic2025} further
couples a visual policy that generates keypoint commands with a low-level
tracker for loco-manipulation. These systems establish motion tracking as an
interface between human motion and robot control. Our focus is a complementary
generative component: an egocentric human motion prior that uses text prompts
to produce SMPL-based motion references for an existing humanoid controller.

\textbf{Vision-language-action models for humanoid robots.}
Vision-language-action models such as OpenVLA~\cite{openvla2024} learn
language-conditioned manipulation policies from robot demonstrations. EgoVLA~\cite{egovla2025}
uses egocentric human videos to learn wrist and hand action prediction, then
retargets these actions and fine-tunes with robot demonstrations for humanoid
manipulation. Related egocentric imitation work, EgoMI~\cite{egomi2025}, models
coordinated head and hand motion for active vision and manipulation on
semi-humanoid embodiments. EgoVLA and EgoMI illustrate how egocentric human
data can support robot policy learning. \method{} instead models human
full-body motion in SMPL space, with visual and text conditioning, and uses the
generated sequence as a reference for a separate whole-body controller.

\textbf{Egocentric human motion understanding and generation.}
Project Aria~\cite{aria2023} provides a platform for multimodal egocentric
sensing. Nymeria~\cite{nymeria2024} pairs egocentric observations with full-body
motion and language annotations; Ego-Exo4D~\cite{grauman2024egoexo4d} captures
skilled activities from synchronized first- and third-person views; and
EgoBody~\cite{egobody2022} provides 3D body annotations during social
interactions. For images captured by a downward-facing camera,
EgoHMR~\cite{egohmr2023} recovers human meshes using a hierarchical latent
diffusion model that combines global and local image features without
requiring 2D pose supervision. This setting provides direct visual evidence
of the wearer's body. Forward-facing observations instead motivate the use
of head-motion cues to infer body motion that lies outside the camera view.
EgoEgo~\cite{li2023egoego} reconstructs full-body motion through
an intermediate head-pose estimate, while EgoAllo~\cite{yi2024estimatingbodyhandmotion}
estimates body and hand motion in the scene coordinate frame from egocentric
SLAM poses and images. UniEgoMotion~\cite{akada2024uniegomotion} unifies
reconstruction, forecasting, and generation using egocentric visual and
trajectory conditions. Building on its task-conditioning formulation,
\method{} introduces a dedicated text stream alongside motion and vision,
using a Triple-stream DiT to learn language-controllable SMPL-based motion
references for humanoid execution. The architecture preserves
modality-specific transformations during joint attention and then fuses the
tokens in Single-stream DiT blocks, with global text conditioning retained
throughout both stages.

\textbf{Language-guided motion generation.}
Text-to-motion methods learn variational motion representations~\cite{guo2022humanml3d,
petrovich2022temos}, denoise motion or latent sequences~\cite{tevet2023mdm,
zhang2022motiondiffuse, mld2023}, or model discrete motion tokens~\cite{chen2023t2mgpt,
motiongpt2023}. TMR~\cite{tmr2023} learns contrastive text--motion embeddings
for retrieval, with motion synthesis as an auxiliary training objective.
These methods establish semantic control and alignment between language and
motion, without directly conditioning on egocentric visual observations.
Beyond text-only conditioning, HUMANISE~\cite{humanise2022} aligns captured
motions with 3D indoor scenes and annotates the resulting interactions with
language. Its scene-and-language conditioned model generates motions that
interact with specified objects, connecting action semantics to scene
geometry. This formulation uses an explicit 3D scene representation to
provide the environmental context for generation.
GMD~\cite{gmd2023} introduces spatial guidance into text-conditioned diffusion
to support trajectory constraints, sparse keyframe locations, and obstacle
avoidance. OmniControl~\cite{omnicontrol2024} extends spatial control to
different joints at arbitrary times, combining spatial guidance with
realism guidance to balance constraint satisfaction and motion coherence.
These approaches expose geometric control through explicit spatial
constraints, alongside the semantic control provided by text.
\method{} uses egocentric image features to supply visual context and
accommodates observed head trajectories through task-dependent masks.
For generation, it conditions on an initial egocentric image and a text
prompt without requiring a prescribed future trajectory. Its multimodal
design therefore addresses how visual observations and language jointly
condition motion when future pose cues are unavailable.
In robotics, language feedback can also correct plans and improve
trajectories~\cite{sharma2022feedback, yell2024, yang2024feedback}.
\method{} uses a prompt selected before generation to steer SMPL-based motion
under egocentric visual conditions; online language correction is outside the
evaluated setting. Its focus is the joint use of scene context and textual
intent in a human motion prior that interfaces with humanoid control.

\section{Method}
\label{sec:method}

\subsection{Overview}
\label{sec:method_overview}

\begin{figure*}[t]
    \centering
    \includegraphics[width=\linewidth]{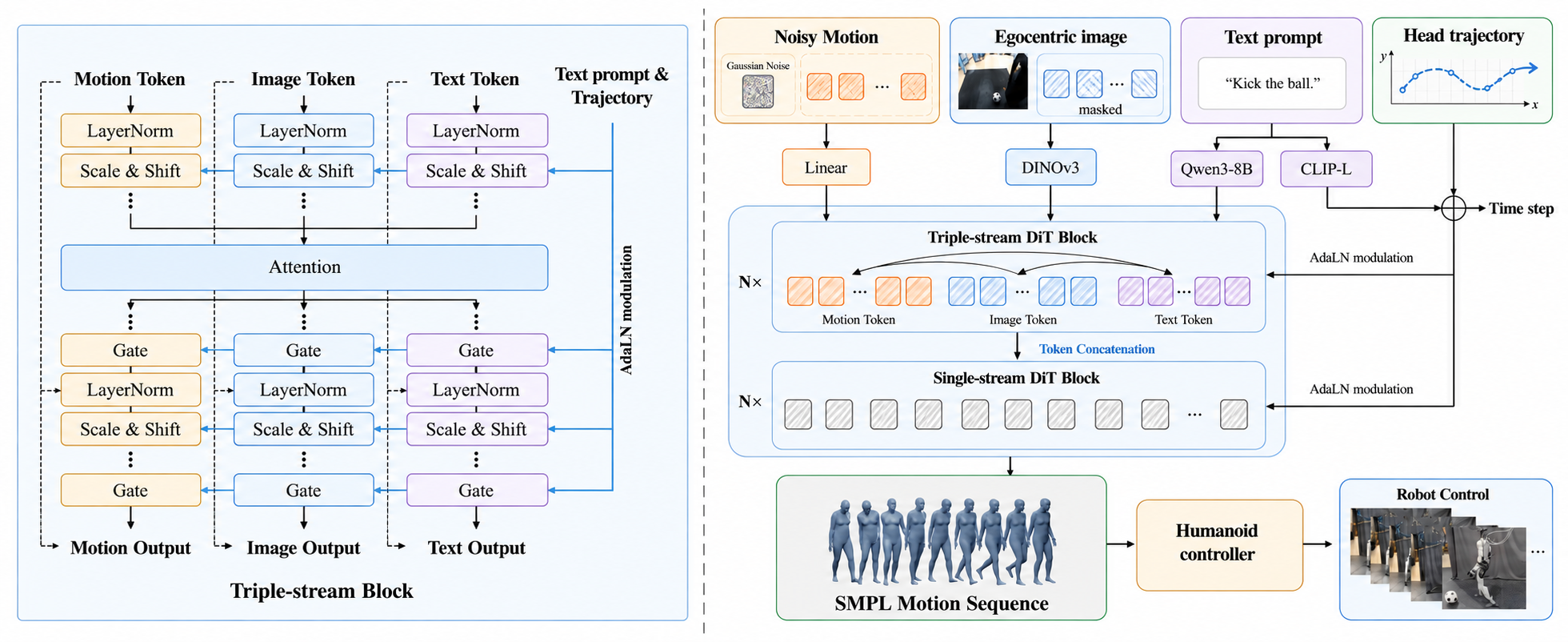}
    \caption{
Overview of \method{}. Right: the framework generates SMPL-based motion conditioned on
egocentric observations, text prompts, and available head trajectories.
Task-dependent visibility masks enable reconstruction, generation, and
forecasting within the same model. The generated motion can directly drive
a humanoid robot through an existing whole-body controller.
Left: a detailed view of a Triple-stream block, showing modality-specific
transformations, joint attention over motion, image, and text tokens,
and global conditioning through AdaLN.
}
    \label{fig:pipeline}
\end{figure*}

\method{} unifies egocentric motion reconstruction, generation, and
forecasting in a text-conditioned model. These tasks
share the same model and differ in their visible conditions:
full-sequence observations for reconstruction, an initial egocentric
image for generation, and past observations for forecasting.
As shown in Figure~\ref{fig:pipeline}, a Triple-stream DiT jointly
processes motion, image, and text tokens to generate SMPL-based
 motion sequences. The generated motions can directly drive
a humanoid robot through an existing whole-body controller.

\subsection{Task Formulation and Flow-Matching Objective}
\label{sec:flow_objective}

We represent the target motion as $\vx_1 \in \mathbb{R}^{T_{\max} \times F}$,
where $T_{\max}$ is the maximum sequence length and $F=234$.
For body motion, we use the rotations and positions of 22 SMPL-X body
joints~\cite{pavlakos2019smplx}, together with foot-contact indicators and
body-shape coefficients. Joint rotations and positions
are expressed in a canonical sequence coordinate frame, with rotations in
continuous 6D form~\cite{zhou2019rotation}. Motion features are standardized
before training.
The condition $\vy$ comprises egocentric image features, head trajectories,
text features, and sequence length, together with visibility and padding masks.
Visibility masks specify which conditions are available, while padding masks
identify valid frames. We learn a conditional velocity field
$v_\theta(\vx_\tau,\tau,\vy)$ using flow matching~\cite{lipman2023flow}.
We sample $\vx_0\sim\mathcal{N}(0,I)$ and $\tau\sim\mathcal{U}(0,1)$, and
linearly interpolate between noise and the target motion. For a single
sequence, the interpolated state, target velocity, and weighted loss are
\begin{equation}
\begin{aligned}
    \vx_{\tau} &= (1-\tau)\vx_0 + \tau\vx_1, \quad
    \mathbf{u}_{\tau}=\vx_1-\vx_0, \\
    \mathcal{L}
    &=
    \frac{
    \sum_{t=1}^{T_{\max}} w_t p_t
    \left\|v_\theta(\vx_\tau,\tau,\vy)_t-\mathbf{u}_{\tau,t}\right\|_2^2
    }{
    F\sum_{t=1}^{T_{\max}} w_t p_t + \epsilon
    } .
\end{aligned}
\label{eq:loss}
\end{equation}
Here $t$ indexes motion frames, $p_t\in\{0,1\}$ excludes padding, and
$\epsilon=10^{-8}$ stabilizes the normalization. The task-dependent frame
weight $w_t$ is 1 for reconstruction and for frames with masked image
conditions. Frames with visible image conditions receive weights of 0.1 for
generation and 0.5 for forecasting, so their motion targets also contribute
to training. The numerator and denominator are accumulated over the minibatch.
All three tasks share the same velocity field and objective, with differences
specified by condition visibility and frame weights.

\subsection{Triple-stream DiT for Egocentric Motion Generation}
\label{sec:triple_stream_dit}

Egocentric motion generation involves three heterogeneous token types with
different roles. The noisy motion tokens represent the body state being
denoised, the image tokens provide egocentric scene and interaction context,
and the text tokens encode high-level semantic guidance. Directly concatenating
these tokens at the input mixes modalities with different temporal structures
and semantic meanings from the beginning. Instead, \method{} adopts a
Triple-stream DiT that first preserves modality-specific processing and then
performs cross-modal fusion.

Each stream has its own input projection, QKV projection,
output projection, and MLP. This design allows each modality to maintain its own
representation space before fusion: motion tokens focus on kinematic
continuity, image tokens focus on scene-grounded cues, and text tokens focus on
prompt semantics. Missing or hidden conditions are replaced
by learned mask tokens, enabling the same architecture to handle different token
combinations.
The three streams exchange information through joint attention. In each
Triple-stream block, we compute stream-specific queries, keys, and values for
motion tokens, image tokens, and text tokens, and then concatenate them to form
a joint attention sequence. This allows motion tokens to query image tokens and
text tokens, while still using stream-specific transformations before and after
attention. Motion and image tokens use RoPE~\cite{su2021rope} to encode
temporal order, whereas text tokens keep the positional structure inherited from
the language encoder. CLIP-L provides a sentence-level embedding as a global
text condition, complementing the text tokens used in joint attention.
Combined with embeddings of the flow timestep, head trajectory, and sequence
length, it forms the conditioning vector for AdaLN-Zero in every Triple-stream
and Single-stream block. This global text conditioning acts across all token
positions and remains active after the streams are merged.
After the Triple-stream stage, motion, image, and text tokens are
concatenated and processed by Single-stream DiT blocks. This stage further
fuses motion, visual, and textual information through shared attention and
MLP layers, while global text conditioning via AdaLN-Zero provides persistent
semantic guidance to align the generated motion with the text prompt.
The resulting motion tokens are projected to the velocity field $v_\theta$.

\subsection{Unified Task Modeling and Sampling}
\label{sec:task_conditioning}

Task-dependent visibility masks specify which modalities and time intervals
condition the shared model~\cite{akada2024uniegomotion}.
Reconstruction uses egocentric images and head trajectories over the full
sequence. Generation retains only the initial egocentric image and masks all
head-trajectory conditions. Forecasting uses images and head trajectories
from an observed prefix, with future observations masked. Text prompts,
when available, condition all three tasks.

Heterogeneous training samples provide complementary supervision for the
shared motion model. Text--motion pairs associate action semantics with
motion patterns, while egocentric video--motion pairs relate visual context
to body movement. Samples containing all three modalities provide joint
visual and language conditioning. Modality availability depends on the
annotations, whereas task-dependent visibility determines which available
conditions the model can access. An unannotated modality and a condition
hidden by the selected task are both handled through masking.

For all three tasks, sampling starts from Gaussian noise
$\vx_0\sim\mathcal{N}(0,I)$. We integrate the learned velocity field
$v_\theta(\vx_\tau,\tau,\vy)$ from $\tau=0$ to $\tau=1$ using Euler steps,
with the selected conditions held fixed. The resulting motion features
are denormalized and decoded into SMPL-based motion sequences.

\subsection{From Generated Motion to Humanoid Control}
\label{sec:humanoid_control}

\method{} connects egocentric observations and text prompts to
humanoid execution through SMPL-based motion references.
The generated motion sequence is converted into the input
representation of an existing whole-body controller, which tracks
the reference to produce robot actions. In this pipeline,
\method{} generates motion conditioned on scene context and user
intent, while the controller handles robot-specific motion tracking.
The motion generator is trained independently of the controller,
without controller or simulator feedback. We evaluate the resulting
pipeline on a Unitree humanoid in
Section~\ref{subsec:humanoid_execution}.

\raggedbottom
\section{Experiments}
\label{sec:experiments}

\subsection{Experimental Setup}
\label{subsec:setup}

\noindent\textbf{Datasets and tasks.}
We train \method{} on a mixture of Nymeria~\cite{nymeria2024},
HumanML3D~\cite{guo2022humanml3d}, and processed
EgoExo4D~\cite{grauman2024egoexo4d}, with respective sampling probabilities
of 70\%, 15\%, and 15\%.
Training clips contain up to 240 frames at 30\,fps in our SMPL-based
representation. Generation is
evaluated on Nymeria and the UniEgoMotion-compatible EgoExo4D validation
split; reconstruction and forecasting are evaluated on Nymeria.

\noindent\textbf{Baselines and metrics.}
We compare reconstruction with EgoAllo~\cite{yi2024estimatingbodyhandmotion},
EgoEgo~\cite{li2023egoego}, and UniEgoMotion~\cite{akada2024uniegomotion}.
For generation and forecasting, we retrain
UniEgoMotion on Nymeria and processed
EgoExo4D without text conditioning. Both methods use the same evaluation
clips and SMPL conversion, while HumanML3D is included only in the
training of \method{}.

We report joint and root-trajectory errors in meters. For reconstruction,
MPJPE measures mean per-joint position error after a single similarity
alignment over the entire sequence, while PA-MPJPE uses per-frame
Procrustes alignment. Root Mean and Root Final measure the average and
final root-position errors after matching the pelvis positions at the
first valid frame with a single translation.
For generation and forecasting, we compute M-FID as the Fr\'echet distance
between generated and reference motion distributions in the
TMR~\cite{tmr2023} feature space. SemSim measures similarity between
paired generated and reference motion embeddings. Air ratio and foot
sliding quantify contact-related artifacts. Lower is better except for SemSim.

\noindent\textbf{Implementation.}
The DiT has a hidden dimension of 1024, 16 attention heads,
16 Triple-stream blocks, and 16 Single-stream blocks.
We train with AdamW, a learning rate of
$5{\times}10^{-5}$, weight decay 0.01, and gradient clipping at 1.0.
Reconstruction, generation, and forecasting are sampled with probabilities
40\%, 30\%, and 30\%, respectively.
Training takes approximately 72 hours on 24 NVIDIA A100 GPUs.
Inference uses 50-step Euler sampling.

\subsection{Egocentric Motion Reconstruction}
\label{subsec:reconstruction}

Reconstruction evaluates how accurately \method{} recovers body pose and
root motion from full-sequence egocentric observations.
Table~\ref{tab:reconstruction_common} compares methods on 3,000 Nymeria
sequences at 15\,fps using 21 common joints.
Scores are computed on jointly valid frames and averaged over sequences.
EgoAllo~\cite{yi2024estimatingbodyhandmotion},
EgoEgo~\cite{li2023egoego} receive GT-derived camera
trajectories.
\method{} and UniEgoMotion~\cite{akada2024uniegomotion} is evaluated without external
conditioning for this comparison.

\begin{table}[H]
    \centering
    \footnotesize
    \setlength{\tabcolsep}{1.2pt}
    \renewcommand{\arraystretch}{1.10}
    \caption{Motion reconstruction on Nymeria.}
    \label{tab:reconstruction_common}
    \begin{tabular*}{\columnwidth}{@{\extracolsep{\fill}}ccccc@{}}
        \toprule
        Method & MPJPE$\downarrow$ & PA-MPJPE$\downarrow$
        & Root Mean$\downarrow$ & Root Final$\downarrow$ \\
        \midrule
        EgoAllo & 0.1885 & 0.1232 & 0.5348 & 0.9444 \\
        EgoEgo & 0.1923 & 0.1332 & 0.5042 & 0.8857 \\
        UniEgoMotion & 0.3890 & 0.1114 & 1.0626 & 1.8229 \\
        \midrule
        \method{} & \textbf{0.1760} & \textbf{0.0754} & \textbf{0.4260} & \textbf{0.8104} \\
        \bottomrule
    \end{tabular*}
\end{table}

\method{} achieves the lowest errors across all four metrics in
Table~\ref{tab:reconstruction_common}, reducing MPJPE and PA-MPJPE by
6.6\% and 32.3\% relative to the best baseline for each metric.
Root Mean decreases by 15.5\% relative to EgoEgo, while the lower
Root Final indicates more accurate recovery of the trajectory endpoint.
Together, these gains demonstrate improved estimation of both body pose
and root motion within our unified framework. Triple-stream DiT retains
modality-specific processing in a dedicated image stream, while joint
attention allows motion tokens to query egocentric visual cues and text
semantics before further fusion in Single-stream blocks. The generation
ablation in Section~\ref{subsec:ablation} also supports this design:
Triple-stream DiT outperforms the Single-stream baseline that concatenates
all modality tokens at the backbone input.

\subsection{Egocentric Motion Generation and Forecasting}
\label{subsec:egocentric_generation}

\noindent\textbf{Motion generation.}
Given an initial egocentric image and available text prompts, \method{}
generates an SMPL-based motion sequence from noise without observed motion
frames or head-trajectory conditioning. We compare with UniEgoMotion on
Nymeria and EgoExo4D in Table~\ref{tab:generation_comparison}.
Since these sparse conditions admit multiple plausible motions, we assess
agreement with reference motions, motion distribution quality, and
foot-contact behavior.

\method{} improves all six metrics over UniEgoMotion on both datasets.
On Nymeria, MPJPE and foot sliding decrease by 59.2\% and 71.2\%,
respectively, indicating gains in motion accuracy and contact behavior.
On EgoExo4D, MPJPE decreases by 7.4\%, accompanied by lower PA-MPJPE,
air ratio, and foot sliding. Lower M-FID and higher SemSim on both datasets
further indicate closer agreement with the reference motion distributions
and greater semantic similarity to individual reference motions.
Together, these results support motion generation from sparse visual
conditions within our unified framework. Triple-stream DiT combines
initial scene cues and prompt semantics through joint attention, while
global text conditioning provides semantic guidance throughout denoising.

\begin{table}[H]
    \centering
    \footnotesize
    \setlength{\tabcolsep}{0.8pt}
    \renewcommand{\arraystretch}{1.90}
    \caption{Egocentric motion generation on Nymeria and EgoExo4D.}
    \label{tab:generation_comparison}
    \begin{tabular*}{\columnwidth}{@{\extracolsep{\fill}}cccccccc@{}}
        \toprule
        & Method & MPJPE$\downarrow$
        & PA-MPJPE$\downarrow$
        & Air$\downarrow$ & M-FID$\downarrow$ & SemSim$\uparrow$ & Slide$\downarrow$ \\
        \midrule
        \multirow{2}{*}{\rotatebox[origin=c]{90}{\scriptsize EgoExo4D}}
        & UniEgoMotion & 0.488 & 0.116 & 0.055 & 0.041 & 0.836 & 1.84 \\
        & \method{} & \textbf{0.452} & \textbf{0.106} & \textbf{0.041} & \textbf{0.035} & \textbf{0.852} & \textbf{1.42} \\
        \midrule
        \multirow{2}{*}{\rotatebox[origin=c]{90}{\scriptsize Nymeria}}
        & UniEgoMotion & 1.168 & 0.127 & 0.091 & 0.086 & 0.689 & 4.37 \\
        & \method{} & \textbf{0.477} & \textbf{0.113} & \textbf{0.024} & \textbf{0.032} & \textbf{0.855} & \textbf{1.26} \\
        \bottomrule
    \end{tabular*}
\end{table}

\begin{figure*}[!t]
    \centering
    \includegraphics[width=\linewidth]{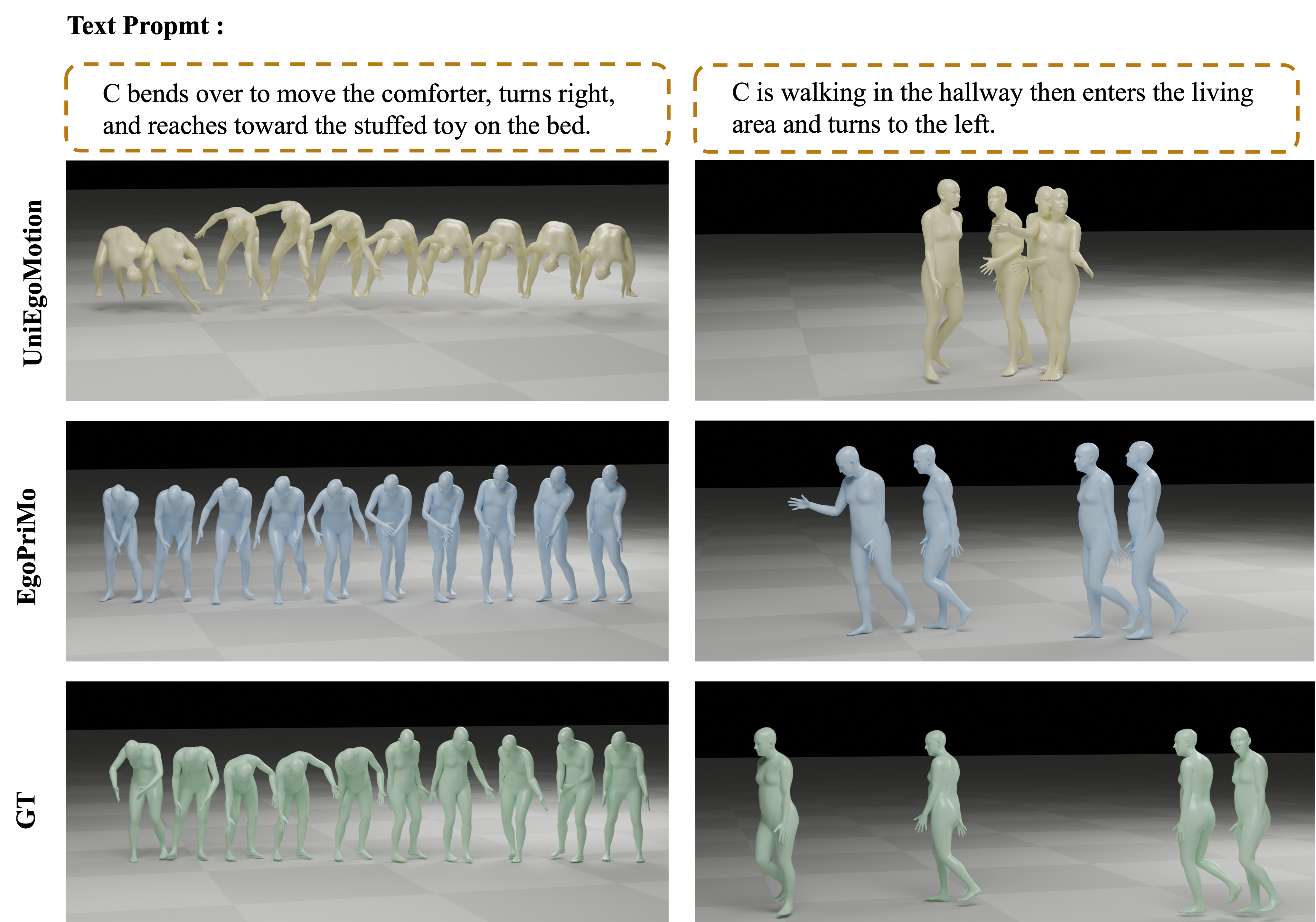}
    \caption{Motion generation examples for bending and reaching on the left
    and walking and turning on the right. Rows show UniEgoMotion, \method{},
    and ground-truth motions; text prompts are shown above.}
    \label{fig:generation_qualitative}
\end{figure*}

Figure~\ref{fig:generation_qualitative} compares generated and reference
motions for bending and reaching, and for walking and turning. In the
first example, \method{} recovers the lowered torso and forward-reaching
pose seen in the reference. The second example illustrates changes in body
orientation during locomotion. These examples illustrate the recovered posture and orientation changes;
the quantitative evaluation covers the broader validation sets.

\noindent\textbf{Prompt control.}
\label{subsec:prompt_control}
To examine the role of language in motion generation, we test whether
changing the text prompt shifts generated motion toward the corresponding
reference action under fixed visual evidence. The study uses
100 paired Nymeria validation clips from 51 recording/action scenes.
Prompt A describes the target clip; a different Prompt B and its reference
motion are selected from another scene with the same clip length. We replace
both token- and sentence-level text features while fixing all other inputs,
task masks, model weights, and initial noise. Sampling uses the final
16-Triple/16-Single checkpoint, 50 Euler steps, no classifier-free guidance,
and base seed 62 with identical noise for each prompt pair.

For each prompt, Table~\ref{tab:prompt_control} compares the generated
motion against both reference motions. SemSim is
$(1+\cos(e_M,e_R))/2$, where $e_M$ and $e_R$ are the TMR embeddings of the
generated and reference motions. Features are evaluated at 20\,fps after
resampling the 30\,fps sequences. Each cell reports a mean over the 100
pairs; the score measures motion-semantic similarity to a prompt-associated
reference.

\begin{table}[H]
    \centering
    \footnotesize
    \caption{Prompt-conditioned motion similarity.}
    \label{tab:prompt_control}
    \begin{tabular}{lcc}
        \toprule
        Generated with & Reference A & Reference B \\
        \midrule
        Prompt A & \textbf{0.8267} & 0.6453 \\
        Prompt B & 0.6789 & \textbf{0.8066} \\
        \bottomrule
    \end{tabular}
\end{table}

The mean scores favor the prompt-associated reference in both conditions:
0.8267 versus 0.6453 for Prompt A, and 0.8066 versus 0.6789 for Prompt B.
The mean matched-minus-mismatched reference
margins are 0.1815 and 0.1277. Their paired-bootstrap 95\% intervals are
[0.1586, 0.2038] and [0.1015, 0.1532], using 20,000 resamples of the 100
pairs. The change in mean reference preference supports prompt-dependent
shifts in motion semantics under fixed visual inputs and sampling noise.
These aggregate scores do not establish instruction compliance for every
clip. 

\noindent\textbf{Motion forecasting.}
\label{subsec:unified_task}
Forecasting tests whether the shared model can extend a partially observed
motion into the future. We keep past motion or observations visible and
predict the future interval using the forecasting mask, without changing
model weights or adding a task-specific head.
Table~\ref{tab:unified_task} reports the Nymeria results.

\begin{table}[H]
    \centering
    \footnotesize
    \setlength{\tabcolsep}{1.5pt}
    \renewcommand{\arraystretch}{1.10}
    \caption{Motion forecasting on Nymeria.}
    \label{tab:unified_task}
    \begin{tabular*}{\columnwidth}{@{\extracolsep{\fill}}ccccccc@{}}
        \toprule
        Method & \shortstack{MPJPE$\downarrow$}
        & \shortstack{PA-MPJPE$\downarrow$}
        & Air$\downarrow$ & M-FID$\downarrow$ & SemSim$\uparrow$ & Slide$\downarrow$ \\
        \midrule
        UniEgoMotion & 0.888 & 0.124 & 0.079 & 0.066 & 0.801 & 2.72 \\
        \method{} & \textbf{0.456} & \textbf{0.111} & \textbf{0.026} & \textbf{0.029} & \textbf{0.860} & \textbf{1.24} \\
        \bottomrule
    \end{tabular*}
\end{table}

\method{} reduces forecasting MPJPE by 48.6\%, from 0.888 to 0.456\,m,
and improves all five remaining metrics. In particular, SemSim increases
from 0.801 to 0.860 and foot sliding decreases from 2.72 to 1.24.
The consistent gains across geometry, motion-semantic similarity, and
contact metrics support forecasting with the generation checkpoint under
this protocol.

\begin{figure*}[!t]
    \centering
    \begin{minipage}[t]{0.48\textwidth}
        \centering
        \includegraphics[width=\linewidth]{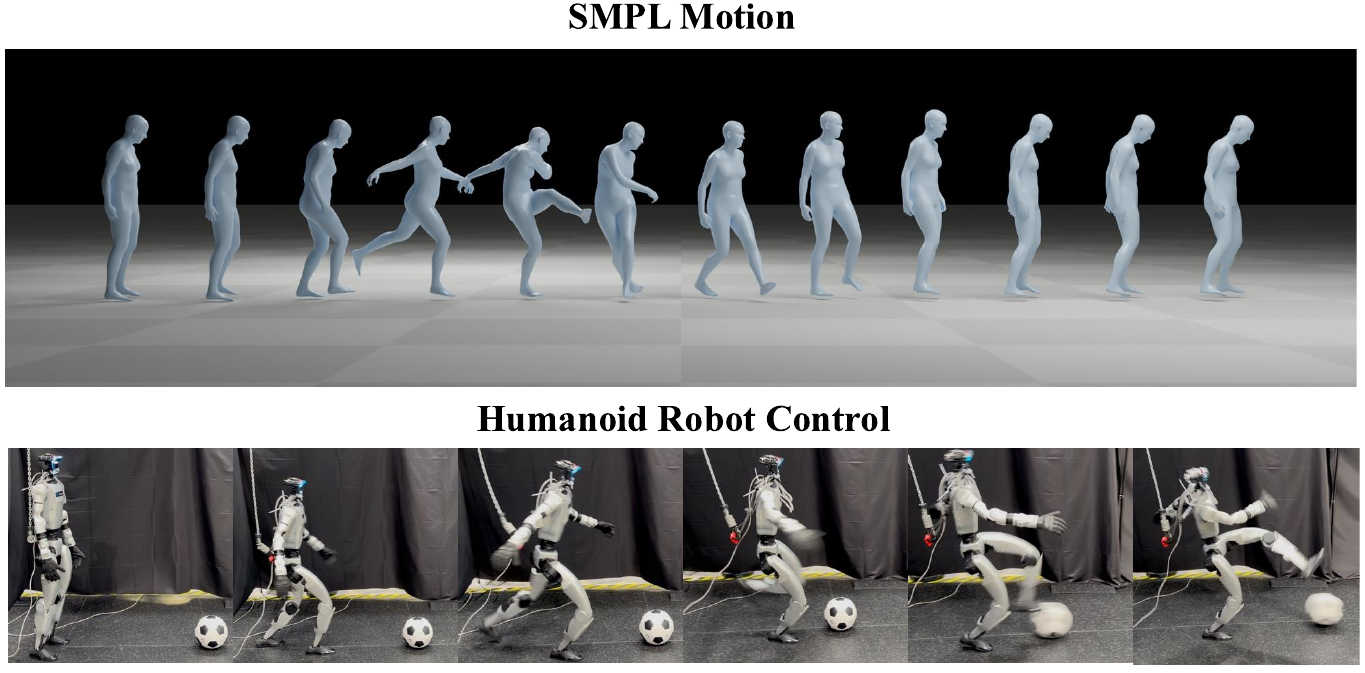}\\
        \small (a) Kicking a soccer ball with the left foot.
    \end{minipage}\hfill
    \begin{minipage}[t]{0.48\textwidth}
        \centering
        \includegraphics[width=\linewidth]{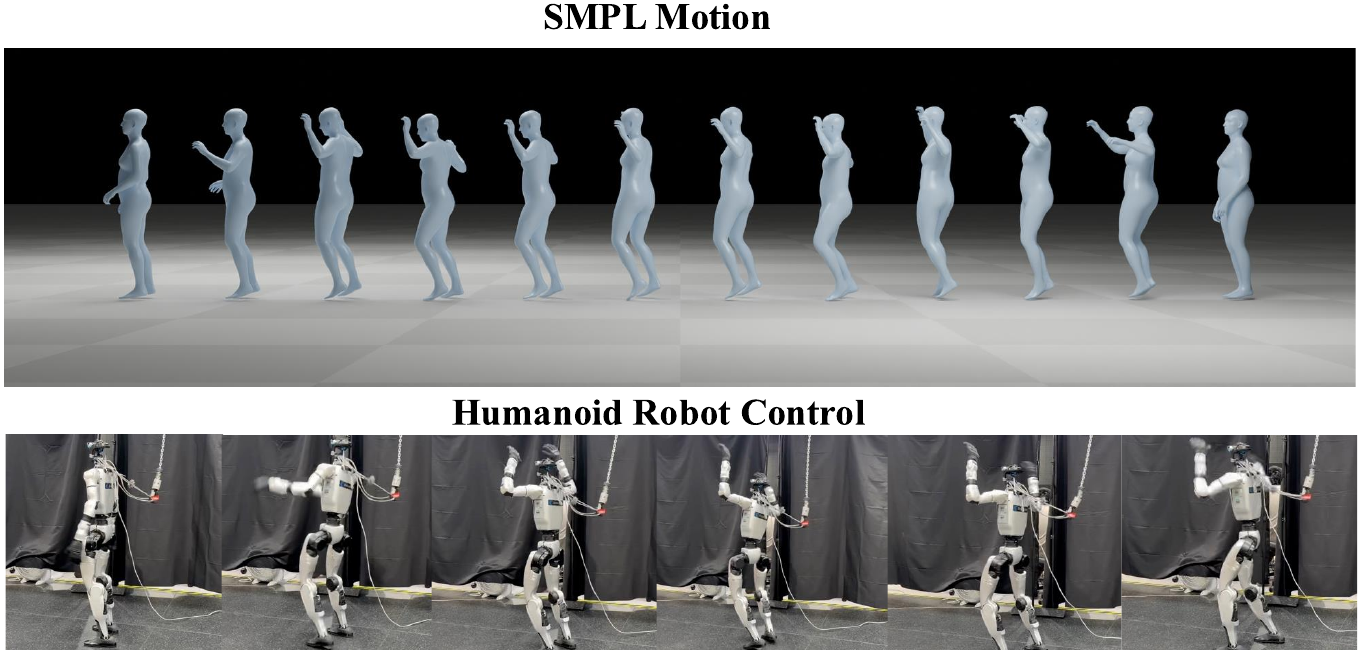}\\
        \small (b) Raising both hands and dancing.
    \end{minipage}
    \caption{Real-robot execution of generated motions for ball kicking on the left
    and dancing on the right.}
    \label{fig:humanoid_execution}
\end{figure*}

\subsection{Ablation Study}
\label{subsec:ablation}

Table~\ref{tab:ablation} evaluates the training data mixture and fusion
architecture under the same Nymeria generation protocol. The data ablation
compares Nymeria-only training with the full mixture while keeping
Triple-stream DiT fixed. The architecture ablation uses the full mixture
for both models; the Single-stream baseline concatenates motion, image,
and text tokens at the backbone input.

\begin{table}[H]
    \centering
    \footnotesize
    \setlength{\tabcolsep}{2pt}
    \renewcommand{\arraystretch}{1.08}
    \caption{Data and fusion ablations.}
    \label{tab:ablation}
    \begin{tabular*}{\columnwidth}{@{\extracolsep{\fill}}cccc@{}}
        \toprule
        Architecture & \shortstack{Single-stream}
        & \shortstack{Triple-stream}
        & \shortstack{Triple-stream} \\
        Data / Recipe & Full mixture & Nymeria-only & Full mixture \\
        \midrule
        MPJPE$\downarrow$ & 0.497 & 0.560 & \textbf{0.477} \\
        PA-MPJPE$\downarrow$ & 0.118 & \textbf{0.109} & 0.113 \\
        Air$\downarrow$ & 0.038 & 0.045 & \textbf{0.024} \\
        Slide$\downarrow$ & 1.39 & 3.92 & \textbf{1.26} \\
        \bottomrule
    \end{tabular*}
\end{table}


\noindent\textbf{Training data.}
The full mixture reduces MPJPE by 14.8\% and foot sliding by 67.9\%
relative to Nymeria-only training, while also lowering air ratio.
PA-MPJPE is 0.113\,m, compared with 0.109\,m for Nymeria-only training.
These results support using heterogeneous motion data within the shared
training objective, with the clearest gains in overall joint-position
accuracy and foot-contact behavior.

\noindent\textbf{Fusion architecture.}
Triple-stream DiT improves all four metrics over the Single-stream
baseline, reducing MPJPE by 4.0\% and air ratio by 36.8\%.
The joint gains in pose and contact metrics support the staged fusion
design: modality-specific transformations are retained as motion, image,
and text tokens exchange information through joint attention, followed
by further fusion in Single-stream DiT blocks. These findings apply to
the tested configurations; isolating the effect of fusion structure
requires matched parameter counts and repeated training runs.

\subsection{Humanoid Robot Control}
\label{subsec:humanoid_execution}

We evaluate motions generated by \method{} from egocentric observations
and text prompts on a Unitree humanoid using an existing whole-body
controller. The controller tracks the generated SMPL references without
returning execution feedback to the motion generator.
Figure~\ref{fig:humanoid_execution} shows execution for ``Kick the ball
with your left foot'' and ``Put up your hands and dance''.
Robot-camera observations are absent from generator training, and the
demonstrations use no robot-view fine-tuning or domain adaptation.

\noindent\textbf{Task success.}
We quantify task completion separately with 10 trials each for ball
kicking and table avoidance.
The avoidance task uses the prompt ``Walk forward to avoid the table''.
A successful trial requires effective ball contact for kicking or
collision-free route completion for avoidance, with no fall in either
task. Table~\ref{tab:robot_trials} summarizes the outcomes.

\begin{table}[H]
    \centering
    \footnotesize
    \caption{Real-robot task success.}
    \label{tab:robot_trials}
    \begin{tabular}{lcc}
        \toprule
        Task & Successful trials & Success rate \\
        \midrule
        Ball kicking & 7/10 & 70\% \\
        Table avoidance & 8/10 & 80\% \\
        \bottomrule
    \end{tabular}
\end{table}

\begin{figure}[H]
    \centering
    \includegraphics[width=\columnwidth]{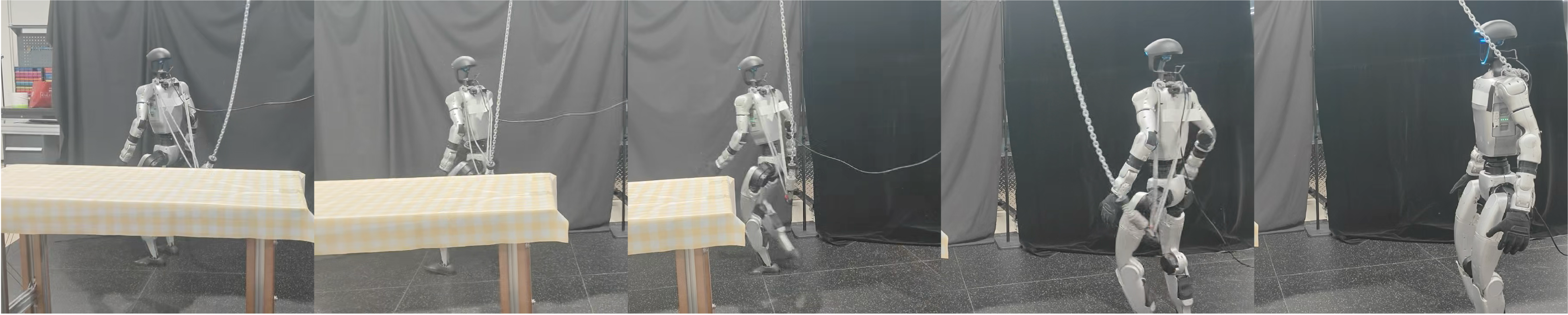}
    \caption{Table avoidance with generated SMPL motion.}
    \label{fig:table_avoidance}
\end{figure}

The pipeline succeeds in 7 of 10 ball-kicking trials and 8 of 10
table-avoidance trials. These results demonstrate that the generated SMPL
references can support object interaction and obstacle avoidance through
an existing controller in the tested setting. The success rates measure
complete task execution across perception, motion generation, and
tracking on this platform. Figure~\ref{fig:table_avoidance} shows an example
of table avoidance.

\section{Discussion and Conclusion}
\label{sec:conclusion}

\method{} unifies egocentric observations, text, and motion in a Triple-stream
DiT. Task-dependent masks enable one checkpoint to support reconstruction,
generation, and forecasting. Experiments demonstrate improved pose accuracy
under the evaluated configurations and prompt-sensitive motion generation.
The generated SMPL references can also be executed through an existing
humanoid controller.

Foot sliding remains a limitation, and the generator does not enforce robot
dynamics or use controller feedback. Real-robot evidence is limited to one
platform and a small number of trials; broader generalization and comparisons
under matched inputs remain to be established. Recording-level training
overlap remains unverified for the reconstruction evaluation.
Future work will focus on
contact-aware generation, execution feedback, and evaluation across datasets
and robot platforms.


\bibliographystyle{IEEEtran}
\bibliography{example}  

@inproceedings{lipman2023flow,
  title={Flow Matching for Generative Modeling},
  author={Lipman, Yaron and Chen, Ricky T. Q. and Ben-Hamu, Heli and Nickel, Maximilian and Le, Matt},
  booktitle={International Conference on Learning Representations},
  year={2023}
}

@inproceedings{tevet2023mdm,
  title={Human Motion Diffusion Model},
  author={Tevet, Guy and Raab, Sigal and Gordon, Brian and Shafir, Yonatan and Cohen-Or, Daniel and Bermano, Amit H.},
  booktitle={International Conference on Learning Representations},
  year={2023}
}

@article{zhang2022motiondiffuse,
  title={MotionDiffuse: Text-Driven Human Motion Generation with Diffusion Model},
  author={Zhang, Mingyuan and Cai, Zhongang and Pan, Liang and Hong, Fangzhou and Guo, Xinying and Yang, Lei and Liu, Ziwei},
  journal={arXiv preprint arXiv:2208.15001},
  year={2022}
}

@inproceedings{chen2023t2mgpt,
  title={Generating Human Motion from Textual Descriptions with Discrete Representations},
  author={Zhang, Jianrong and Zhang, Yangsong and Cun, Xiaodong and Zhang, Yong and Zhao, Hongwei and Lu, Hongtao and Shen, Xi and Shan, Ying},
  booktitle={IEEE/CVF Conference on Computer Vision and Pattern Recognition},
  year={2023}
}

@inproceedings{guo2022humanml3d,
  title={Generating Diverse and Natural 3D Human Motions from Text},
  author={Guo, Chuan and Zou, Shihao and Zuo, Xinxin and Wang, Sen and Ji, Wei and Li, Xingyu and Cheng, Li},
  booktitle={IEEE/CVF Conference on Computer Vision and Pattern Recognition},
  year={2022}
}

@inproceedings{petrovich2022temos,
  title={{TEMOS}: Generating Diverse Human Motions from Textual Descriptions},
  author={Petrovich, Mathis and Black, Michael J. and Varol, Gul},
  booktitle={European Conference on Computer Vision},
  year={2022}
}

@inproceedings{grauman2024egoexo4d,
  title={{Ego-Exo4D}: Understanding Skilled Human Activity from First- and Third-Person Perspectives},
  author={Grauman, Kristen and Westbury, Andrew and Torresani, Lorenzo and Kitani, Kris and Malik, Jitendra and Afouras, Triantafyllos and Ashutosh, Kumar and Baiyya, Vijay and Bansal, Siddhant and Boote, Bikram and others},
  booktitle={IEEE/CVF Conference on Computer Vision and Pattern Recognition},
  pages={19383--19400},
  year={2024}
}

@inproceedings{li2023egoego,
  title={Ego-Body Pose Estimation via Ego-Head Pose Estimation},
  author={Li, Jiaman and Liu, Karen and Wu, Jiajun},
  booktitle={IEEE/CVF Conference on Computer Vision and Pattern Recognition},
  pages={17142--17151},
  year={2023}
}

@inproceedings{akada2024uniegomotion,
  title={UniEgoMotion: A Unified Model for Egocentric Motion Reconstruction, Forecasting, and Generation},
  author={Patel, Chaitanya and Nakamura, Hiroki and Kyuragi, Yuta and Kozuka, Kazuki and Niebles, Juan Carlos and Adeli, Ehsan},
  booktitle={IEEE/CVF International Conference on Computer Vision},
  year={2025},
  note={arXiv:2508.01126},
}

@inproceedings{pavlakos2019smplx,
  title={Expressive Body Capture: 3D Hands, Face, and Body from a Single Image},
  author={Pavlakos, Georgios and Choutas, Vasileios and Ghorbani, Nima and Bolkart, Timo and Osman, Ahmed A. A. and Tzionas, Dimitrios and Black, Michael J.},
  booktitle={IEEE/CVF Conference on Computer Vision and Pattern Recognition},
  year={2019}
}

@inproceedings{zhou2019rotation,
  title={On the Continuity of Rotation Representations in Neural Networks},
  author={Zhou, Yi and Barnes, Connelly and Lu, Jingwan and Yang, Jimei and Li, Hao},
  booktitle={Proceedings of the IEEE/CVF Conference on Computer Vision and Pattern Recognition},
  pages={5745--5753},
  year={2019}
}

@article{su2021rope,
  title={RoFormer: Enhanced Transformer with Rotary Position Embedding},
  author={Su, Jianlin and Lu, Yu and Pan, Shengfeng and Murtadha, Ahmed and Wen, Bo and Liu, Yunfeng},
  journal={arXiv preprint arXiv:2104.09864},
  year={2021}
}

@inproceedings{nymeria2024,
  title={{Nymeria}: A Massive Collection of Multimodal Egocentric Daily Motion in the Wild},
  author={Ma, Lingni and Ye, Yuting and Hong, Fangzhou and Guzov, Vladimir and Jiang, Yifeng and Postyeni, Rowan and Pesqueira, Luis and Gamino, Alexander and Baiyya, Vijay and Kim, Hyo Jin and Bailey, Kevin and Fosas, David S. and Liu, C. Karen and Liu, Ziwei and Engel, Jakob and De Nardi, Renzo and Newcombe, Richard},
  booktitle={European Conference on Computer Vision},
  pages={445--465},
  doi={10.1007/978-3-031-72691-0_25},
  year={2024},
}

@article{aria2023,
  title={Project Aria: A New Tool for Egocentric Multi-Modal AI Research},
  author={Engel, Jakob and Somasundaram, Kiran and Goesele, Michael and Sun, Albert and Gamino, Alexander and Turner, Andrew and Talattof, Arjang and Yuan, Arnie and Souti, Bilal and Meredith, Brighid and others},
  journal={arXiv preprint arXiv:2308.13561},
  year={2023}
}

@inproceedings{egobody2022,
  title={EgoBody: Human Body Shape and Motion of Interacting People from Head-Mounted Devices},
  author={Zhang, Siwei and Ma, Qianli and Zhang, Yan and Qian, Zhiyin and Kwon, Taein and Pollefeys, Marc and Bogo, Federica and Tang, Siyu},
  booktitle={European Conference on Computer Vision},
  year={2022}
}

@inproceedings{mld2023,
  title={Executing your Commands via Motion Diffusion in Latent Space},
  author={Chen, Xin and Jiang, Biao and Liu, Wen and Huang, Zilong and Fu, Bin and Chen, Tao and Yu, Gang},
  booktitle={IEEE/CVF Conference on Computer Vision and Pattern Recognition},
  year={2023}
}

@inproceedings{tmr2023,
  title={TMR: Text-to-Motion Retrieval Using Contrastive 3D Human Motion Synthesis},
  author={Petrovich, Mathis and Black, Michael J. and Varol, Gul},
  booktitle={IEEE/CVF International Conference on Computer Vision},
  year={2023}
}

@article{motiongpt2023,
  title={MotionGPT: Human Motion as a Foreign Language},
  author={Jiang, Biao and Chen, Xin and Liu, Wen and Yu, Jingyi and Yu, Gang and Chen, Tao},
  journal={arXiv preprint arXiv:2306.14795},
  year={2023}
}

@article{humanplus2024,
  title={HumanPlus: Humanoid Shadowing and Imitation from Humans},
  author={Fu, Zipeng and Zhao, Qingqing and Wu, Qi and Wetzstein, Gordon and Finn, Chelsea},
  journal={arXiv preprint arXiv:2406.10454},
  year={2024}
}

@inproceedings{omnih2o2024,
  title={OmniH2O: Universal and Dexterous Human-to-Humanoid Whole-Body Teleoperation and Learning},
  author={He, Tairan and Luo, Zhengyi and He, Xialin and Xiao, Wenli and Zhang, Chong and Zhang, Weinan and Kitani, Kris and Liu, Changliu and Shi, Guanya},
  booktitle={Conference on Robot Learning},
  year={2024},
  note={arXiv:2406.08858},
}

@article{exbody2024,
  title={Expressive Whole-Body Control for Humanoid Robots},
  author={Cheng, Xuxin and Ji, Yandong and Chen, Junming and Yang, Ruihan and Yang, Ge and Wang, Xiaolong},
  journal={arXiv preprint arXiv:2402.16796},
  year={2024}
}

@article{hover2024,
  title={HOVER: Versatile Neural Whole-Body Controller for Humanoid Robots},
  author={He, Tairan and Xiao, Wenli and Lin, Toru and Luo, Zhengyi and Xu, Zhenjia and Jiang, Zhenyu and Kautz, Jan and Liu, Changliu and Shi, Guanya and Wang, Xiaolong and Fan, Linxi and Zhu, Yuke},
  journal={arXiv preprint arXiv:2410.21229},
  year={2024}
}

@inproceedings{asap2025,
  title={ASAP: Aligning Simulation and Real-World Physics for Learning Agile Humanoid Whole-Body Skills},
  author={He, Tairan and Gao, Jiawei and Xiao, Wenli and Zhang, Yuanhang and Wang, Zi and Wang, Jiashun and Luo, Zhengyi and He, Guanqi and Sobanbabu, Nikhil and Pan, Chaoyi and Yi, Zeji and Qu, Guannan and Kitani, Kris and Hodgins, Jessica K. and Fan, Linxi and Zhu, Yuke and Liu, Changliu and Shi, Guanya},
  booktitle={Robotics: Science and Systems},
  year={2025},
  note={arXiv:2502.01143},
}

@article{visualmimic2025,
  title={VisualMimic: Visual Humanoid Loco-Manipulation via Motion Tracking and Generation},
  author={Yin, Shaofeng and Ze, Yanjie and Yu, Hong-Xing and Liu, C. Karen and Wu, Jiajun},
  journal={arXiv preprint arXiv:2509.20322},
  year={2025}
}

@article{sonic2025,
  title={{SONIC}: Supersizing Motion Tracking for Natural Humanoid Whole-Body Control},
  author={Luo, Zhengyi and Yuan, Ye and Wang, Tingwu and Li, Chenran and Casta{\~n}eda, Fernando and Chen, Sirui and Cao, Zi-Ang and Li, Jiefeng and Minor, David and Ben, Qingwei and Park, Jinhyung and Sami, David and Wang, Zi and Da, Xingye and Ding, Runyu and Hogg, Cyrus and Song, Lina and Lim, Edy and Jeong, Eugene and He, Tairan and Xue, Haoru and Xiao, Wenli and Yuen, Simon and Kautz, Jan and Chang, Yan and Iqbal, Umar and Fan, Linxi and Zhu, Yuke},
  journal={Science Robotics},
  volume={11},
  number={117},
  pages={eaed4592},
  doi={10.1126/scirobotics.aed4592},
  year={2026}
}

@inproceedings{sharma2022feedback,
  title={Correcting Robot Plans with Natural Language Feedback},
  author={Sharma, Pratyusha and Sundaralingam, Balakumar and Blukis, Valts and Paxton, Chris and Hermans, Tucker and Torralba, Antonio and Andreas, Jacob and Fox, Dieter},
  booktitle={Robotics: Science and Systems},
  year={2022},
  note={arXiv:2204.05186},
}

@article{yell2024,
  title={Yell at your Robot: Improving On-the-Fly from Language Corrections},
  author={Shi, Lucy Xiaoyang and Hu, Zheyuan and Zhao, Tony Z. and Sharma, Archit and Pertsch, Karl and Luo, Jianlan and Levine, Sergey and Finn, Chelsea},
  journal={arXiv preprint arXiv:2403.12910},
  year={2024}
}

@inproceedings{yang2024feedback,
  title={Trajectory Improvement and Reward Learning from Comparative Language Feedback},
  author={Yang, Zhaojing and Jun, Miru and Tien, Jeremy and Russell, Stuart and Dragan, Anca and B{\i}y{\i}k, Erdem},
  booktitle={Conference on Robot Learning},
  year={2024},
  note={arXiv:2410.06401},
}

@article{openvla2024,
  title={OpenVLA: An Open-Source Vision-Language-Action Model},
  author={Kim, Moo Jin and Pertsch, Karl and Karamcheti, Siddharth and Xiao, Ted and Balakrishna, Ashwin and Nair, Suraj and Rafailov, Rafael and Foster, Ethan and Lam, Grace and Sanketi, Pannag and others},
  journal={arXiv preprint arXiv:2406.09246},
  year={2024}
}

@article{egovla2025,
  title={EgoVLA: Learning Vision-Language-Action Models from Egocentric Human Videos},
  author={Yang, Ruihan and Yu, Qinxi and Wu, Yecheng and Yan, Rui and Li, Borui and Cheng, An-Chieh and Zou, Xueyan and Fang, Yunhao and Cheng, Xuxin and Qiu, Ri-Zhao and Yin, Hongxu and Liu, Sifei and Han, Song and Lu, Yao and Wang, Xiaolong},
  journal={arXiv preprint arXiv:2507.12440},
  year={2025}
}

@article{egomi2025,
  title={EgoMI: Learning Active Vision and Whole-Body Manipulation from Egocentric Human Demonstrations},
  author={Yu, Justin and Shentu, Yide and Wu, Di and Abbeel, Pieter and Goldberg, Ken and Wu, Philipp},
  journal={arXiv preprint arXiv:2511.00153},
  year={2025}
}

@inproceedings{yi2024estimatingbodyhandmotion,
      title={Estimating Body and Hand Motion in an Ego-sensed World}, 
      author={Brent Yi and Vickie Ye and Maya Zheng and Yunqi Li and Lea M{\"u}ller and Georgios Pavlakos and Yi Ma and Jitendra Malik and Angjoo Kanazawa},
      booktitle={IEEE/CVF Conference on Computer Vision and Pattern Recognition},
      pages={7072--7084},
      year={2025},
}

@inproceedings{egohmr2023, title={{EgoHMR}: Egocentric Human Mesh Recovery via Hierarchical Latent Diffusion Model}, DOI={10.1109/icra48891.2023.10161247}, booktitle={2023 IEEE International Conference on Robotics and Automation (ICRA)}, author={Liu, Yuxuan and Yang, Jianxin and Gu, Xiao and Guo, Yao and Yang, Guang-Zhong}, year={2023}, month=May, pages={9807--9813} }

@inproceedings{humanise2022,
 author = {Wang, Zan and Chen, Yixin and Liu, Tengyu and Zhu, Yixin and Liang, Wei and Huang, Siyuan},
 booktitle = {Advances in Neural Information Processing Systems},
 doi = {10.52202/068431-1088},
 pages = {14959--14971},
 title = {{HUMANISE}: Language-conditioned Human Motion Generation in 3D Scenes},
 volume = {35},
 year = {2022}
}

@InProceedings{gmd2023,
    author    = {Karunratanakul, Korrawe and Preechakul, Konpat and Suwajanakorn, Supasorn and Tang, Siyu},
    title     = {Guided Motion Diffusion for Controllable Human Motion Synthesis},
    booktitle = {Proceedings of the IEEE/CVF International Conference on Computer Vision (ICCV)},
    month     = {October},
    year      = {2023},
    pages     = {2151-2162}
}

@inproceedings{omnicontrol2024,
 author = {Xie, Yiming and Jampani, Varun and Zhong, Lei and Sun, Deqing and Jiang, Huaizu},
 booktitle = {International Conference on Learning Representations},
 pages = {28176--28194},
 title = {{OmniControl}: Control Any Joint at Any Time for Human Motion Generation},
 year = {2024}
}


\end{document}